\documentclass[10pt,twocolumn,letterpaper]{article}

\usepackage[algorithms]{wacv} 

\usepackage{graphicx} 

\usepackage{booktabs}
\usepackage{tabularx}
\usepackage{amsmath,amssymb,amsfonts}
\usepackage{algpseudocode}
\usepackage{algorithm}
\usepackage{array}
\usepackage{uniquecounter}
\usepackage{textcomp}
\usepackage{stfloats}
\usepackage{url}
\usepackage{verbatim}
\usepackage{graphicx}
\usepackage{cite}
\usepackage{xcolor}
\usepackage{subcaption}
\usepackage{xspace}
\usepackage{units}
\usepackage{diagbox}
\usepackage{multirow}
\usepackage{bbm}
\usepackage{amsmath}

\definecolor{wacvblue}{rgb}{0.21,0.49,0.74}
\usepackage[pagebackref,breaklinks,colorlinks]{hyperref}

\usepackage[capitalize]{cleveref}
\crefname{section}{Sec.}{Secs.}
\Crefname{section}{Section}{Sections}
\Crefname{table}{Table}{Tables}
\crefname{table}{Tab.}{Tabs.}

\def\wacvPaperID{2550} 
\def\confName{WACV}
\def\confYear{2026}

\newcommand{\method}{Attention-based Double Compression\xspace}
\newcommand{\methoda}{ADC\xspace}

\newcommand{\cifar}{CIFAR100\xspace}
\newcommand{\food}{Food101\xspace}

\begin{document}
\title{Communication Efficient Split Learning of ViTs with \\ \method }

\author{Federico Alvetreti \textsuperscript{1} \qquad
Jary Pomponi \textsuperscript{2,3} \footnotemark[1]  \qquad  Paolo Di Lorenzo \textsuperscript{2, 3}   \qquad Simone Scardapane \textsuperscript{2, 3} \\[0.5ex]
\textsuperscript{1} Department of Computer, Control, and Management Engineering (DIAG) \\
\textsuperscript{2} Department of Information Engineering, Electronics, and Telecommunications (DIET) \\ \textsuperscript{3} Consorzio Nazionale Interuniversitario per le Telecomunicazioni (CNIT) \\
}


\maketitle

\iftoggle{wacvfinal}{
\renewcommand{\thefootnote}{\fnsymbol{footnote}}
\footnotetext[1]{Corresponding author: jary.pomponi@uniroma1.it}
\renewcommand{\thefootnote}{\arabic{footnote}}
\setcounter{footnote}{0}
}


\newcommand\blfootnote[1]{
    \begingroup
    \renewcommand\thefootnote{}\footnote{#1}
    \addtocounter{footnote}{-1}
    \endgroup
}

\begin{abstract}
This paper proposes a novel communication-efficient Split Learning (SL) framework, named \method (\methoda), which reduces the communication overhead required for transmitting intermediate Vision Transformers activations during the SL training process. \methoda incorporates two parallel compression strategies. The first one merges samples' activations that are similar, based on the average attention score calculated in the last client layer; this strategy is class-agnostic, meaning that it can also merge samples having different classes, without losing generalization ability nor decreasing final results. The second strategy follows the first and discards the least meaningful tokens, further reducing the communication cost. Combining these strategies not only allows for sending less during the forward pass, but also the gradients are naturally compressed, allowing the whole model to be trained without additional tuning or approximations of the gradients. Simulation results demonstrate that \method outperforms state-of-the-art SL frameworks by significantly reducing communication overheads while maintaining high accuracy.
\end{abstract}

\iftoggle{wacvfinal}{
\blfootnote{This work has been supported by: 1) SNS JU project 6G-GOALS under the EU’s Horizon program Grant Agreement No 101139232; 2) Sapienza grant RG123188B3EF6A80 (CENTS), by European Union under the Italian National Recovery and Resilience Plan of NextGenerationEU, partnership on Telecommunications of the Future (PE00000001 - program RESTART); 3) "Sapienza,  Avvio alla ricerca" grant (UGOV 1201260).  We also acknowledge ISCRA for awarding this project access to the LEONARDO supercomputer, owned by the EuroHPC Joint Undertaking, hosted by CINECA (Italy).}\vspace{-.4cm}
}

\section{Introduction}
\begin{figure}[t]
    \centering
    \includegraphics[width=\linewidth]{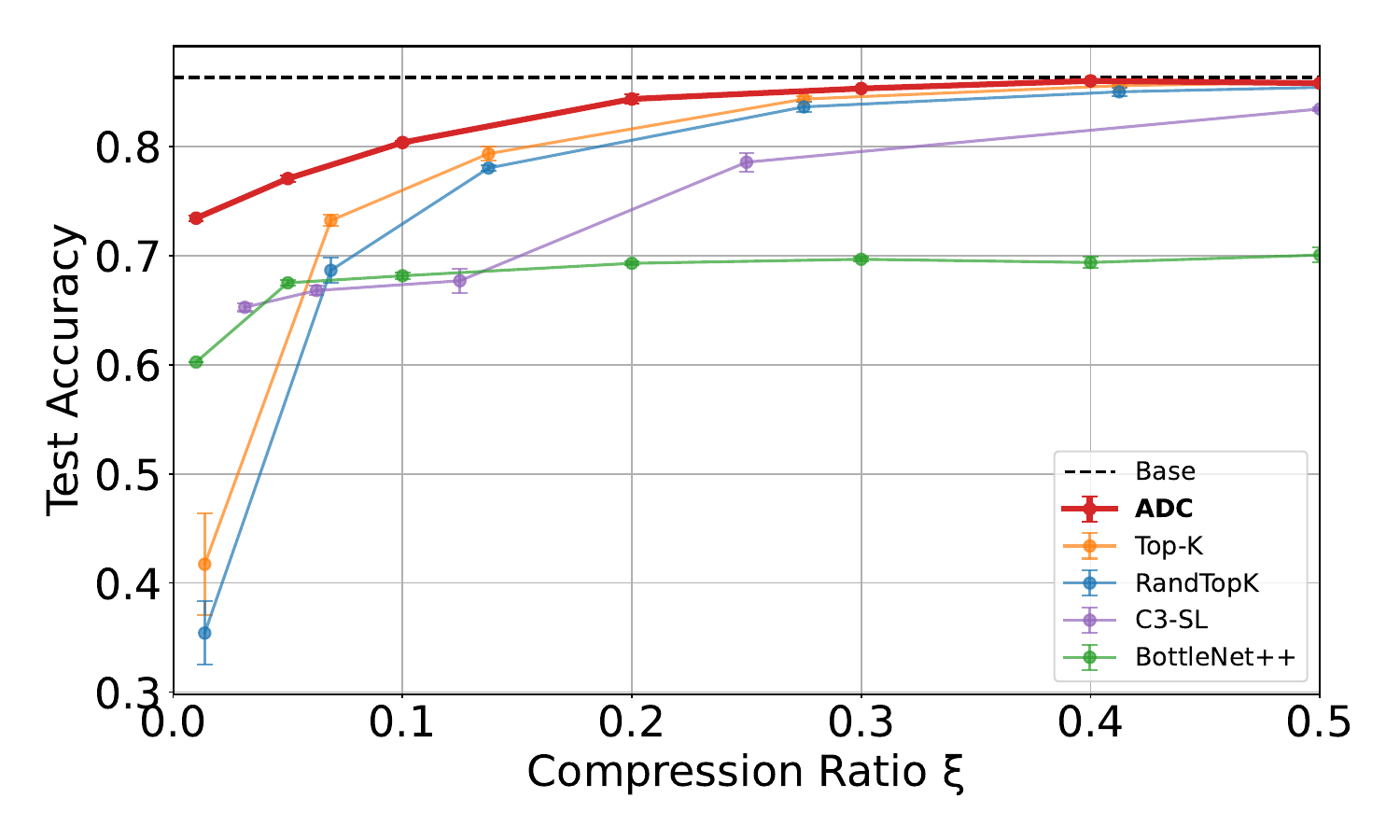}
    \caption{\methoda consistently surpasses other baselines across all compression ratios $\xi$, enabling Split Learning to operate reliably even under extreme communication constraints while incurring only minimal accuracy loss. Results obtained training DeiT-S on \cifar.}
    \label{fig:main_result}
\end{figure}
The proliferation of deep neural networks (DNNs) across diverse domains such as computer vision, natural language processing, and medical diagnostics has revolutionized artificial intelligence applications. However, the computational intensity and memory requirements of DNN training present significant challenges for deployment on resource-constrained edge devices. Traditional cloud-based learning paradigms \cite{CHAN2023126327}, while computationally feasible, necessitate the transmission of raw data from edge devices to centralized cloud servers. This approach not only generates substantial communication overhead but also raises critical privacy concerns regarding sensitive user data.

Split Learning (SL) \cite{singh2025machine, lin2024split, 10.1145/3527155, 10.1145/3037697.3037698} emerges as a promising solution to reconcile the computational demands of DNN training with privacy preservation and communication efficiency. The fundamental principle of SL involves partitioning a neural network architecture between edge devices and cloud infrastructure. Specifically, the client device executes the initial layers ($f_c$) using local data, while the cloud server processes the subsequent layers ($f_s$). The training is performed in three steps. It starts with the forward propagation through $f_c$, followed by the transmission of extracted features and corresponding labels to the server model $f_s$, which continues forward propagation through its layers. Then, it computes the gradients and propagates them backward to the client, which updates its model. This collaborative training framework transmits only intermediate activations and the corresponding gradients. The size of the transmitted data depends on the batch size and the number of features produced by the client. 

Despite these advantages, communication bottlenecks remain a significant limitation in practical SL implementations. To address the challenge of communication overhead, various communication-efficient frameworks have been proposed \cite{9706690, 8645013, 8824955, 8639121, binucci2024enabling, 9145068, hsieh2022c3, zheng2023reducing, 9685493}, with the primary objective of compressing features or gradients during the training procedure. To do that, two representative approaches have been developed: 1) using autoencoders, and 2) applying sparsification or quantization.
In the first approach, an autoencoder is inserted at the output of the device-side mode, and the decoder at the server side \cite{9685493, devoto2025adaptive}. This configuration enables the reduction of the number of symbols for both client-produced features and gradient matrices. The second approach reduces the features by employing compression techniques, e.g., sparsification or quantization, to the intermediate features and/or the gradients \cite{zheng2023reducing, yuan2020federated}. Quantization is a prominent method, aiming to reduce the communication cost by quantizing each entry of output activations \cite{li2023psaq, liu2022distributed, tao2021fat, liu2023noisyquant, wu2020easyquant, yvinec2023powerquant} or grouping similar vectors into clusters and representing them with a shared codebook entry, an approach called vector quantization \cite{oh2023fedvqcs, shlezinger2020uveqfed}.  Other approaches, instead, sparsify the activations or the gradients by transmitting only non-zero, or the most significant, values \cite{9660377, 10327766}. The approaches in the two sets are orthogonal, and more than one can be employed at the same time to boost the compression.

Despite advancements in communication-efficient Split Learning systems, these approaches continue to struggle with maintaining model accuracy while reducing data transmission overhead. The core issue with current methodologies lies in their one-size-fits-all compression strategy, where all feature representations and gradient updates receive identical treatment regardless of their significance to the learning process. This indiscriminate approach results in valuable information being compressed at the same rate as less crucial data, leading to deteriorated model quality, especially when aggressive compression ratios are employed. Consequently, there is a critical need for intelligent, context-aware compression mechanisms that can dynamically adjust based on the relative importance of different data components in Split Learning environments, thereby achieving better bandwidth utilization while preserving training effectiveness.

\textbf{Contribution}: this paper presents a novel communication-efficient SL framework, named \method (\methoda), which reduces the communication overhead of SL while maintaining high performance. The core idea of \method is to leverage Transformer-based model properties to compress transmitted features in two steps: firstly, similar batch samples are combined; secondly, the resulting tokens are compressed. Compared to other approaches, our proposal compresses the batch in an unsupervised way by looking at the output activations. Then, once compressed, features' number is further diminished by keeping only the most important tokens. Through extensive numerical evaluation of various image classification tasks, we demonstrate the superiority of the proposed approach. \Cref{fig:main_result} anticipates such results by showing that our approach achieves the best test accuracy for all possible compression ratios of the transmitted symbols.


\section{Background on Vision Transformers}

In this section, we provide a brief overview of the Vision Transformer (ViT) model \cite{dosovitskiy2021an}, which serves as a core component of our proposal.


A ViT model $f(\mathbf{x})$ takes an image $\mathbf{x} \in \mathbb{R}^{C\cdot H \cdot W}$ as input, where $C$, $H$, and $W$ represent the number of channels, height, and width of the image, respectively. The overall structure of the model is defined as follows:
\begin{equation}\label{eq:VIT}
    f(\mathbf{x}) = \mathcal{C} \circ \mathcal{B}^L \circ \mathcal{B}^{L-1} \circ \cdots \circ \mathcal{B}^{1} \circ \mathcal{E}(\mathbf{x})
\end{equation}
where $\mathcal{E}(\mathbf{x})$ is a preprocessing layer that turns the image into a sequence $\mathbf{H}^0 \in \mathbb{R}^{n \times d}$, with $n$ the number of tokens and $d$ their length, for a total size of $D = n \cdot d$ features; then, $\{\mathcal{B}^1 \dots \mathcal{B}^{L}\}$ are transformer blocks that process the tokens via multi-head attention (MHA) and feed-forward networks, with $L$ denoting the number of blocks. The set of tokens is created by dividing an image into non-overlapping patches of fixed length, which are then flattened and projected to a fixed embedding size (i.e., $d$) using a trainable network.  To preserve spatial information, a unique learnable positional embedding is added to each token, encoding the original position of each patch within the image. Finally, the tokens set includes a trainable class token, which is used as input to the classification layer $\mathcal{C}$ to produce the final prediction vector. 

The MHA mechanism represents the core of each transformer block. This mechanism consists of $H$ parallel self-attention heads, each computing outputs by analyzing interactions between tokens in the input sequence. These outputs are then concatenated along the feature dimension and projected through a learnable matrix $\mathbf{W}_o \in \mathbb{R}^{Hd_v \times d}$, where $d_v$ is the output dimension of each attention head and $d$ is the model’s hidden dimension. Letting $\mathbf{Z}^{l-1} \in \mathbb{R}^{n \times d}$ denote the sequence of token embeddings input to the $l$-th transformer block, the MHA function reads as:
\begin{equation}\label{eq:MHA}
  \text{MHA}^l(\mathbf{Z}^{l-1}) = \left[\text{SA}^l_1(\mathbf{Z}^{l-1}), \dots, \text{SA}^l_H(\mathbf{Z}^{l-1})\right] \mathbf{W}_o  
\end{equation}

for $l=1,\ldots,L$, where $\text{SA}_i$ is Self-Attention head:
\begin{equation*}\label{eq:SA}
\text{SA}_i^l(\mathbf{Z}^{l-1}) = \underbrace{\text{softmax}\left(\frac{\mathbf{Q}_i \mathbf{K}_i^\top}{\sqrt{d_k}}\right)}_{\text{Attention score }  \mathbf{A}_i^l} \mathbf{V}_i,
\end{equation*}
for $i=1,\ldots,H$, where the softmax is applied row-wise, and $\mathbf{A}_i^l \in \mathbb{R}^{n \times n}$ contains a probability vector, called attention score, for each token against all the others. 
Each attention head in (\ref{eq:SA}) applies learned linear projections to the input feature $\mathbf{Z}^{l-1}$  to produce queries $\mathbf{Q}_i \in \mathbb{R}^{n \times d_k}$, keys $\mathbf{K}_i \in \mathbb{R}^{n \times d_k}$, and values $\mathbf{V}_i \in \mathbb{R}^{n \times d_v}$ matrices.
%
%
where $d_k$ is the dimension of the queries and keys.
After computing all $H$ attention outputs, they are concatenated to form a matrix of shape $(n, Hd_v)$ in \cref{eq:MHA}, which is then linearly projected back to dimension $d$ using $\mathbf{W}_o$. A key observation is that the token count $n$ can be dynamically adjusted without compromising the MHA mechanism.


\section{Problem formulation}

Given a client model $f_c$, a server model $f_s$, and a classification dataset $\mathcal{D}$ with $\mathcal{L}$ labels, we aim to fine-tune both models while keeping the communication cost contained.

To construct the two models, we start from a ViT with $L$ layers and select a splitting point $1 < l < L$. The first $l$ blocks are assigned to the client, while the remaining $L - l$ blocks are assigned to the server:
\begin{align*}
    f_c &= \mathcal{B}^{l} \circ \cdots \circ \mathcal{B}^{2} \circ \mathcal{B}^{1} \circ \mathcal{E}, \\ 
    f_s &= \mathcal{C} \circ\ \mathcal{B}^{L} \circ \cdots \circ \mathcal{B}^{l+2} \circ \mathcal{B}^{l+1}. 
\end{align*}

Inspired by \cite{Arachchige_Camtepe_Sun_2022}, we operate in a Split Learning scenario composed of three steps: 

\begin{enumerate}
    \item Client Forward Pass: given a generic training sample tuple $(x, y)$, the client produces the activation vector on its last layer using its model as $z = f_c(x) \in \mathbb{R}^D$, and sends it to the server, along with the label $y$. We refer to the communication cost of this stage as $\overrightarrow{C}$. 
    \item Server Update: the server treats the activations $z$ as inputs to perform one step of gradient descent on the server-side model $f_s(z)$. In addition, the server also computes the gradient with respect to the input activation $g_s = \nabla_z f_s(z)$ and sends it back to the client. We refer to the cost of sending the gradients back to the client as $\overleftarrow{C}$.
    \item Client Backward Pass: The client computes the gradient with respect to the client-side model using the chain rule $g_c = g_s^{\mathsf{T}} J_{w_\mathrm{C}} f_c(x)$, where $w_c$ is the set of parameters of the client model.     
\end{enumerate}

The communication cost of a complete training step is defined as the sum $C = \overrightarrow{C} + \overleftarrow{C}$.

The algorithm composed of the so-defined three steps is equivalent to a mini-batch stochastic gradient descent (SGD) with a total batch size of B, preserving both performance and iteration complexity.

%


\subsection{Training communication cost}
\begin{figure*}[t!]
    \centering
    \includegraphics[width=0.9\linewidth]{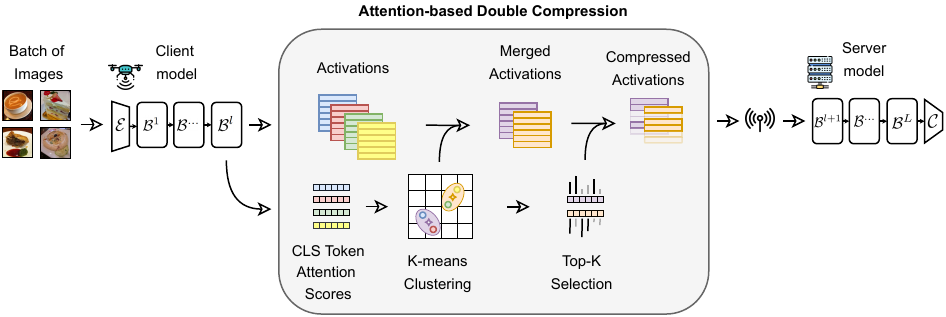}
    \caption{A visualization of the proposed method. The activations of the batch are merged based on cluster similarity. These clusters are computed over the attention scores of the class tokens. Then, of the merged activations, only the most important tokens are kept, and the rest are discarded. The resulting activations are sent to a remote server to complete the training step.}
    \label{fig:diagram}
\end{figure*}
Here, we further assume that the available communication budget is limited, such that only a finite number of symbols can be transmitted over the entire training process. We denote this limit as $\Gamma$, expressed in number of communicable symbols. This setup reflects the constraints of embedded systems, where power and bandwidth are restricted, and provides a common ground for comparing different compression methods.  

The baseline scenario, in which uncompressed activations and gradients are transmitted, is referred to as \textit{base}. Its forward and backward communication costs are defined as:  
\begin{equation*}
    \begin{aligned}
    & \overrightarrow{C}_{\text{base}} = \text{B}(D \phi + \log_2 \mathcal{L}), \\
    & \overleftarrow{C}_{\text{base}} = \text{B}D \phi,
    \end{aligned}
\end{equation*}
where $B$ is the batch size, $D$ is the number of features per sample, and $\phi=32$ represents the cost (in bits) of transmitting a single feature.\footnote{We distinguish symbols by their bit requirements: for instance, a floating-point feature requires 32 bits, whereas a label requires only $\log_2 V$ bits, where $V$ is the maximum label value.}  

For a generic compression method $m$, we define the forward and backward compression ratios, $\overrightarrow{\xi}_m$ and $\overleftarrow{\xi}_m$, as the normalized communication costs relative to the \textit{base} case:  
\[
\overrightarrow{\xi}_m = \frac{\overrightarrow{C}_{m}}{\overrightarrow{C}_{\text{base}}}, 
\quad 
\overleftarrow{\xi}_m = \frac{\overleftarrow{C}_{m}}{\overleftarrow{C}_{\text{base}}}.
\]
The overall compression ratio is then given by the average of the two:  
\[
\xi_m = \frac{\overrightarrow{\xi}_m + \overleftarrow{\xi}_m}{2}.
\]  

Given a method $m$ with compression ratio $\xi_m$, the cumulative communication cost after $i$ training iterations is:  
\[
C_{\text{tot}, m}(i) := i \cdot \xi_m \cdot C_{\text{base}}.
\]
Under the communication constraint $\Gamma$, the maximum number of iterations is therefore bounded as:  
\[
I := \max \{ i \, : \, C_{\text{tot}, m}(i) \le \Gamma \}.
\]  

A lower compression ratio $\xi_m$ reduces the communication cost per iteration, thereby allowing a greater number of training iterations within the same budget. The ultimate goal of a compression method is to achieve this reduction while preserving the performance of the \textit{base} approach.

\section{\method \\ (\methoda)}

\label{sec:proposed}

We propose a two-step compression strategy composed of \emph{batch compression} followed by \emph{token selection}. The first step reduces the number of samples in the batch by merging similar activations, while the second step reduces the dimensionality of the merged activations by retaining only their most relevant tokens.  
By combining both operations, our method compresses along two orthogonal axes --- samples and features --- achieving high compression ratios with minimal performance loss.  
A complete overview is shown in \Cref{fig:diagram}.

\subsection*{1) Batch Compression}
Consider a batch of $B$ samples $\mathbf{X} = \{x_i, y_i\}_{i=1}^B$ and their client-side activations $f_c(\mathbf{X}) \in \mathbb{R}^{B \times D}$.  
Our goal is to reduce the batch dimension $B$ to a target size $T < B$ by merging similar activations.

For a given sample $x$, let $f_c(x) = z \in \mathbb{R}^{n \times d}$ denote its activation matrix.  
To quantify token importance, we exploit the CLS-token attention scores from the last transformer block $\mathcal{B}^l$, averaged across all heads. This produces a vector $\text{CLS}_{score}(z) \in \mathbb{R}^n$, which is already computed during the forward pass and thus requires no extra overhead.  
Prior work \cite{liang2022not, caron2021emerging, chowdhury2025prompt, vilas2023analyzing, walmer2023teaching} has shown that the CLS token consistently attends to task-relevant tokens, making $\text{CLS}_{score}(z)$ a reliable proxy for importance.

We cluster the set of scores $\{\text{CLS}_{score}(z_i)\}_{i=1}^B$ into $T$ groups using $K$-means, yielding centroids $\{\mathbf{C}_i\}_{i=1}^T \in \mathbb{R}^n$.  
Each activation is assigned to its closest centroid:
\begin{equation}
  \mathbbm{1}(z, i) =
  \begin{cases}
    1 & \text{if } i = \arg\min_j \| \text{CLS}_{score}(z) - \mathbf{C}_j \|_2^2, \\
    0 & \text{otherwise}.
  \end{cases}
\end{equation}

The new activations communicated to the server are then computed as cluster averages:
\begin{equation}
  \mathbf{F}_i = \frac{\sum_{j=1}^{B} \mathbbm{1}(z_j, i) \cdot f_c(x_j)}
  {\sum_{j=1}^{B} \mathbbm{1}(z_j, i)}.
\end{equation}

Labels are merged accordingly. Each original label $y_j$ is mapped to its one-hot vector $\text{one-hot}(y_j)$. For cluster $i$, the associated label vector is:
\begin{equation}
  \mathbf{Y}_i = \frac{\sum_{j=1}^{B} \mathbbm{1}(z_j, i) \cdot \text{one-hot}(y_j)}{\sum_{j=1}^{B} \mathbbm{1}(z_j, i)} .
\end{equation}
This yields soft multi-label vectors, where higher values indicate classes more frequently represented in the cluster.  
The final compressed batch is therefore:
\[
  \overline{\mathbf{X}} = \{(\mathbf{F}_i, \mathbf{Y}_i)\}_{i=1}^T.
\]

\subsection*{2) Token Selection}
While batch compression reduces the number of samples, each merged activation $\mathbf{F}_i$ may still contain redundant tokens.  
To further reduce dimensionality, we leverage the same attention information used during merging: for each centroid $\mathbf{C}_i$, we retain only the top-$k$ tokens in $\mathbf{F}_i$ that correspond to the most important positions in $\mathbf{C}_i$.  

This ensures that compression remains consistent across the two phases:  
\begin{itemize}
  \item In \emph{batch compression}, samples are merged because they share similar distributions of important tokens.  
  \item In \emph{token selection}, only those commonly important tokens are preserved, discarding the rest.  
\end{itemize}

By aligning the merging and selection criteria, the method ensures that only the most informative parts of the activations are communicated, leading to efficient compression with minimal impact on downstream accuracy.








\section{Experimental Set-Up}

\iftoggle{wacvfinal}{
In this section, we introduce the experimental setups. The complete code used to run all experiments is available at the following \href{https://github.com/Federico-Alvetreti/Split-Learning}{repository}. 
}

\subsection{Training details}
%
%
We selected and trained two models on two different datasets. The models are the small and tiny versions of DeiT \cite{touvron2021training} (respectively DeiT-S and DeiT-T), and the datasets are \cifar and \food \cite{bossard14}. As the communication budget $\Gamma$, we choose a value that guarantees $10$ epochs of base training, such that $\Gamma = 10 \cdot \lvert \mathcal{D} \rvert \cdot C_{base}$, where $C_{base}$ depends on the model used and $\lvert \mathcal{D} \rvert$ is the size of the training dataset.


For all the experiments, we train until the communication budget $\Gamma$ is reached. We use a batch size of $128$, a split point of $l = 3$, and Adam as optimizer \cite{kingma2014adam}. As the augmentation strategy, we use the same set used in \cite{touvron2021training}.

\subsection{Baselines}
\label{sec:baselines}

We evaluate our method against a diverse set of baselines which are related to our proposal, including both traditional compression methods and state-of-the-art techniques. The baselines are:

\begin{itemize}
    \item \textbf{Base} is the standard training, in which no compression is performed. 

    \item \textbf{BottleNet++}~\cite{eshratifar2019bottlenet} inserts lightweight feed-forward networks immediately before and after the split point, compressing each token’s feature vector from dimensionality $d$ to $d'$, with $d' < d$.

    \item \textbf{Top-K} retains the top $k$ activation values in terms of magnitudes and sets all others to zero. Because the positions of these $k$ elements must be recoverable, their indices also have to be transmitted during the forward pass, introducing an overhead of $\log_2 D/\phi$ bits per symbol.

    \item \textbf{RandTopK}~\cite{zheng2023reducing} extends Top-K by adding a small random perturbation to the feature scores before selection, preventing systematic “starvation” and ensuring that even lower-magnitude activations occasionally get transmitted.

    \item \textbf{C3-SL  \cite{hsieh2022c3}} is a batch compression technique. It takes activations as groups, and  compresses $R$ activations together into a single one using circular convolution. The compressed activations are recovered on the receiver side using superposition. 
\end{itemize}

The forward and backward compression ratios formulas of each baseline, compared to our proposal, are shown in \Cref{tab:compressed-size}.

%
\begin{table}[t]
  \centering
  \scalebox{1}{
  \begin{tabular}{
      l>{$}c<{$}>{$}c<{$}
  }
    \toprule
    \textbf{Method} & \overrightarrow{\xi} & \overleftarrow{\xi}\\
    \midrule
    Base      & 1 & 1 \\
    BottleNet++   & \nicefrac{d'}{d} & \nicefrac{d'}{d} \\
    Top-K         & \nicefrac{k}{D} \left(1 + \log_{2} \nicefrac{D}{\phi} \right) 
                  & \nicefrac{k}{D} 
                  \\
    RandTopK      & \nicefrac{k}{D} \left(1 + \log_{2} \nicefrac{D}{\phi} \right) 
                  & \nicefrac{k}{D} \\ 
    C3-SL         & \nicefrac{1}{R} & \nicefrac{1}{R} \\ 
    \toprule
    \textbf{\methoda}   & \nicefrac{T}{B} \cdot \nicefrac{k}{n} 
                  & \nicefrac{T}{B} \cdot \nicefrac{k}{n} \\
    \bottomrule
  \end{tabular}
  }
  \caption{Compression ratios for all methods. Refer to \Cref{sec:baselines} for each symbols' meaning.}
  \label{tab:compressed-size}
\end{table}
\subsection{Hyperparameters}
\label{sec:hyperparams}
\begin{figure}[t]
    \centering
    \begin{subfigure}[t]{0.32\columnwidth}
        \centering
        \includegraphics[width=1.03\linewidth]{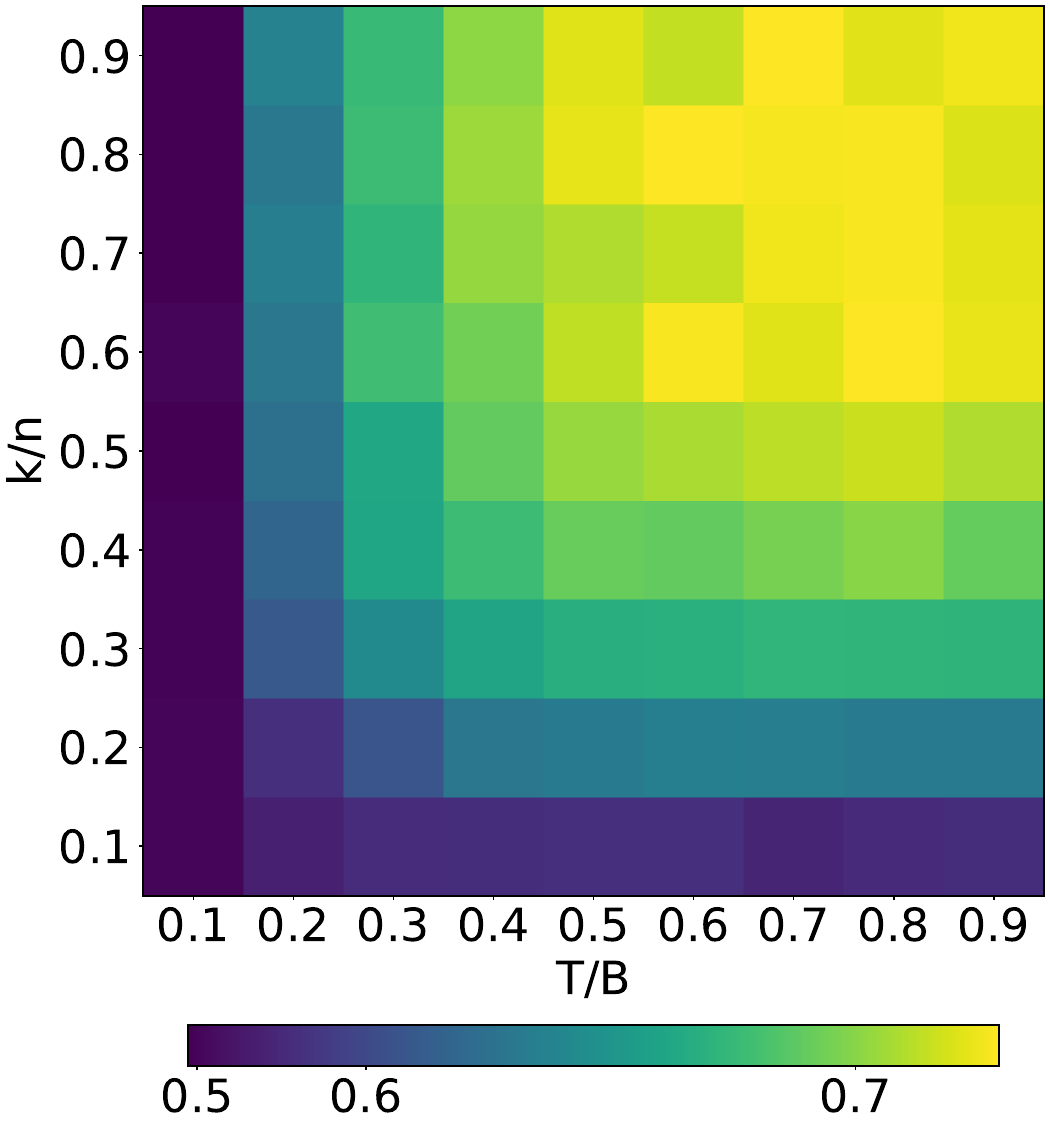}
        \caption{\food.}
        \label{fig:parameter_tuning_food}
    \end{subfigure}
    \hfill
    \begin{subfigure}[t]{0.32\columnwidth}
        \centering
        \includegraphics[width=1.03\linewidth]{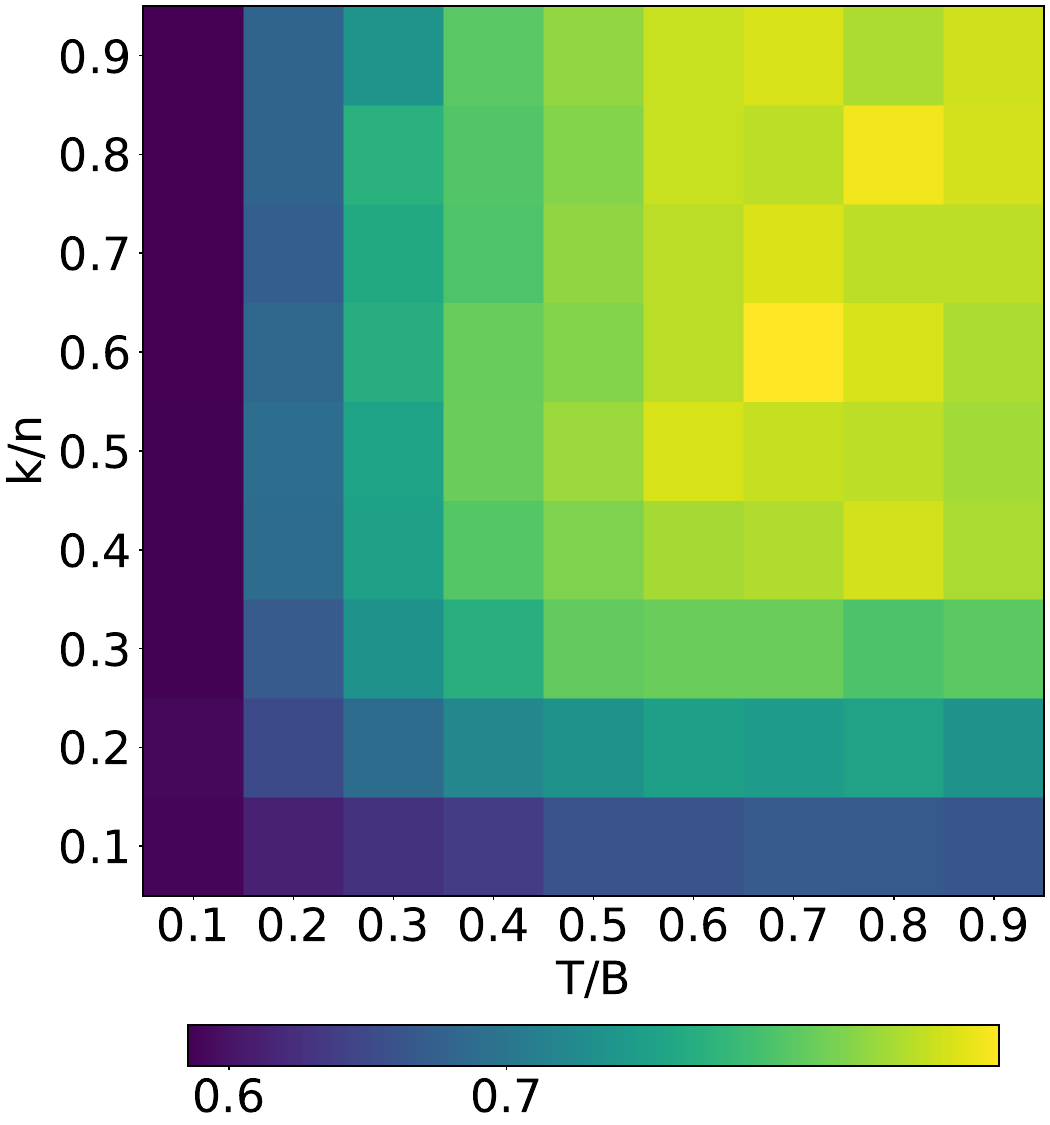}
        \caption{\cifar.}
        \label{fig:parameter_tuning_cifar}
    \end{subfigure}
    \hfill
    \begin{subfigure}[t]{0.33\columnwidth}
        \centering
        \includegraphics[width=\linewidth]{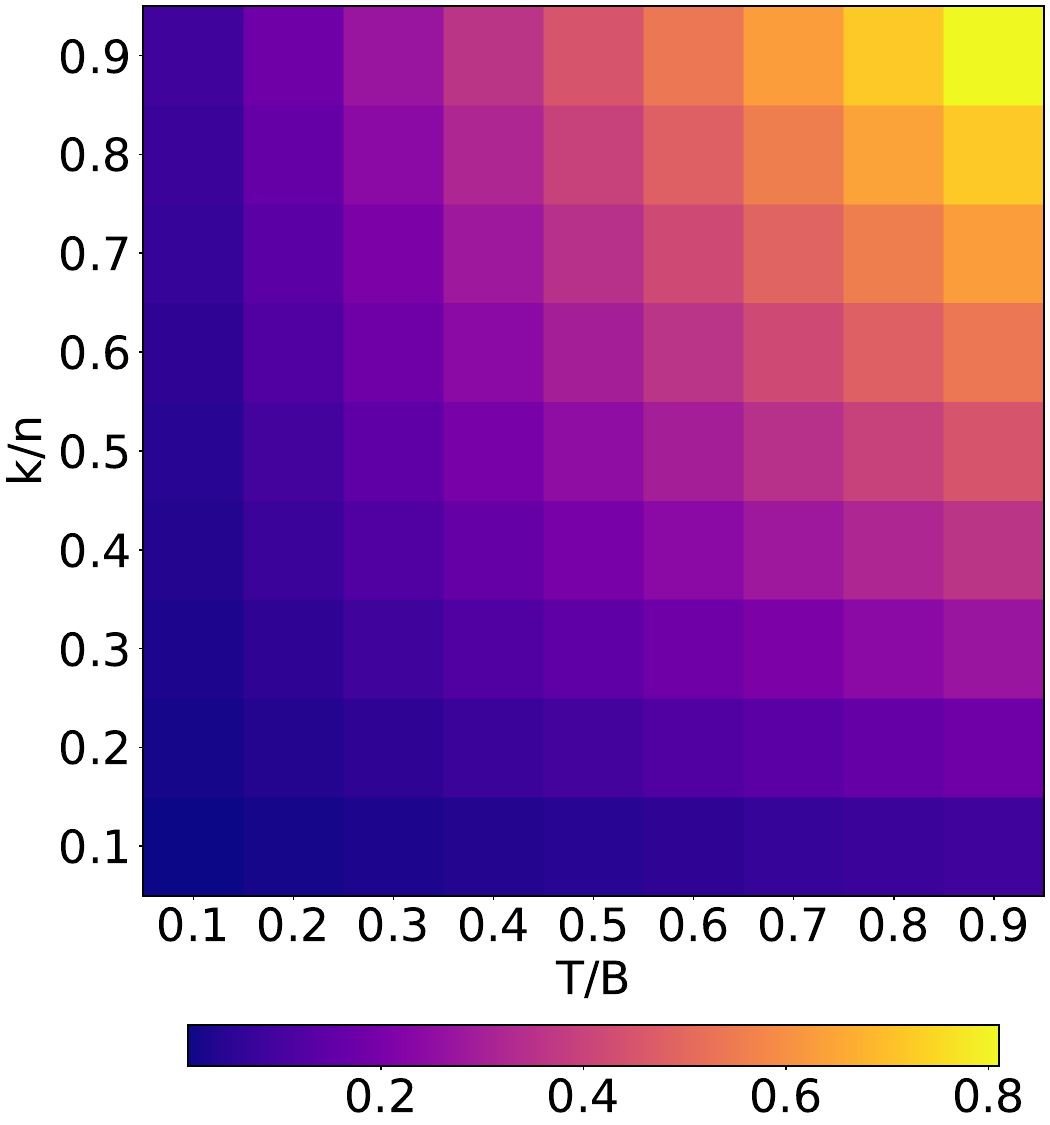}
        \caption{Compression ratio $\xi$.}
        \label{fig:corresponding_ratio}
    \end{subfigure}

    \caption{Effect of $T/B$ and $k/n$ selection on the final validation accuracy for  \food \subref{fig:parameter_tuning_food} and \cifar \subref{fig:parameter_tuning_cifar} when training DeiT-T, and the corresponding overall compression ratio $\xi$ \subref{fig:corresponding_ratio}.}
    \label{fig:parameter_tuning}
\end{figure}
For \textbf{Top-K} and \textbf{RandTopK} we set $k \in [0.01, 0.05, 0.1, 0.2, 0.3, 0.4, 0.5]$. For \textbf{C3-SL} we set the compression factor R to $[2, 4, 8, 16, 32]$. Lastly, for \textbf{BottleNet++} and \textbf{\methoda} we set their respective hyperparameters so that the final compression rate would be $\xi  \in [0.01, 0.05, 0.1, 0.2, 0.3, 0.4, 0.5]$. 

Regarding our proposal, the core intuition is that a trade-off exists between the two compression factors. Here, we validate it by performing a grid search over these two hyperparameters. Each method is evaluated over a randomly extracted development set, not used for training. In \Cref{fig:parameter_tuning}, we evaluate our proposal on the \food \subref{fig:parameter_tuning_food} and \cifar \subref{fig:parameter_tuning_cifar} validation sets using DeiT-T, varying both the number of clusters $T$ and the number of retained top-$k$ tokens after merging. The results show that, when considering the same compression ratio $\xi$ \subref{fig:corresponding_ratio}, the best performance is obtained when the compression factors from batch merging and token selection are equally balanced. In contrast, pushing one of them to the extreme leads to a severe drop in performance. 
A desired compression ratio $\xi \in [0, 1]$ can be obtained by selecting any combination of clusters T and selected top-k tokens such that $\xi = \nicefrac{T}{B} \cdot \nicefrac{k}{n}$
%
%
, and we assume equal contribution from batch merging and token selection by setting $\nicefrac{k}{n} = \nicefrac{T}{B} = \sqrt{\xi}$. This is the formulation we use in all the experiments. 

\section{Experimental results}
\subsection{Main results}
\begin{figure*}[tbp]
    \centering
    \begin{subfigure}{0.48\linewidth}
        \centering
        \includegraphics[width=0.92\linewidth]{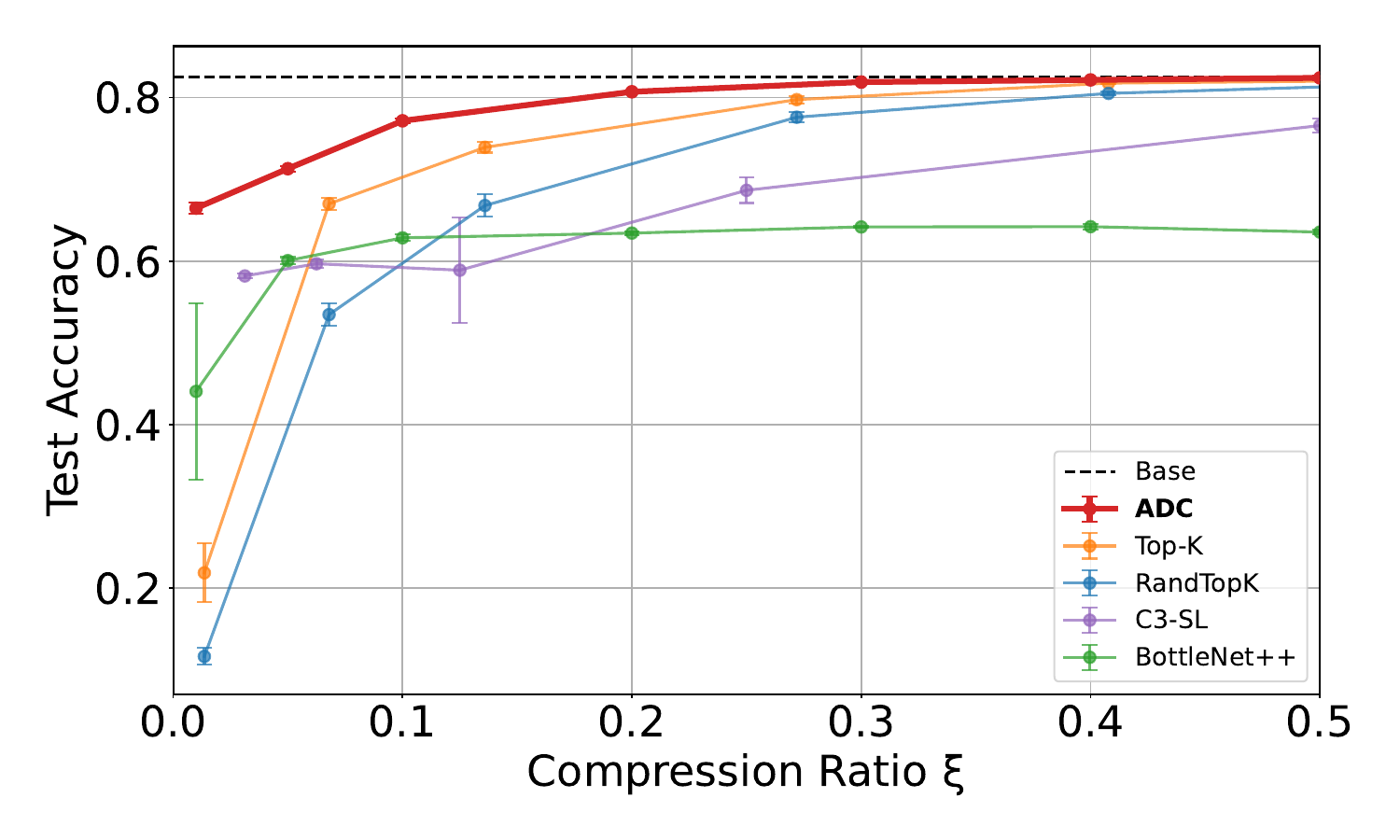}
        \caption{DeiT-T on \cifar}
    \end{subfigure}
    \hfill
    \begin{subfigure}{0.48\linewidth}
        \centering
        \includegraphics[width=0.92\linewidth]{images/results/summary_plots/cifar100_small.pdf}
        \caption{DeiT-S on \cifar}
        \label{fig:sub2}
    \end{subfigure}

    \vspace{0.5em}

    \begin{subfigure}{0.48\linewidth}
        \centering
        \includegraphics[width=0.92\linewidth]{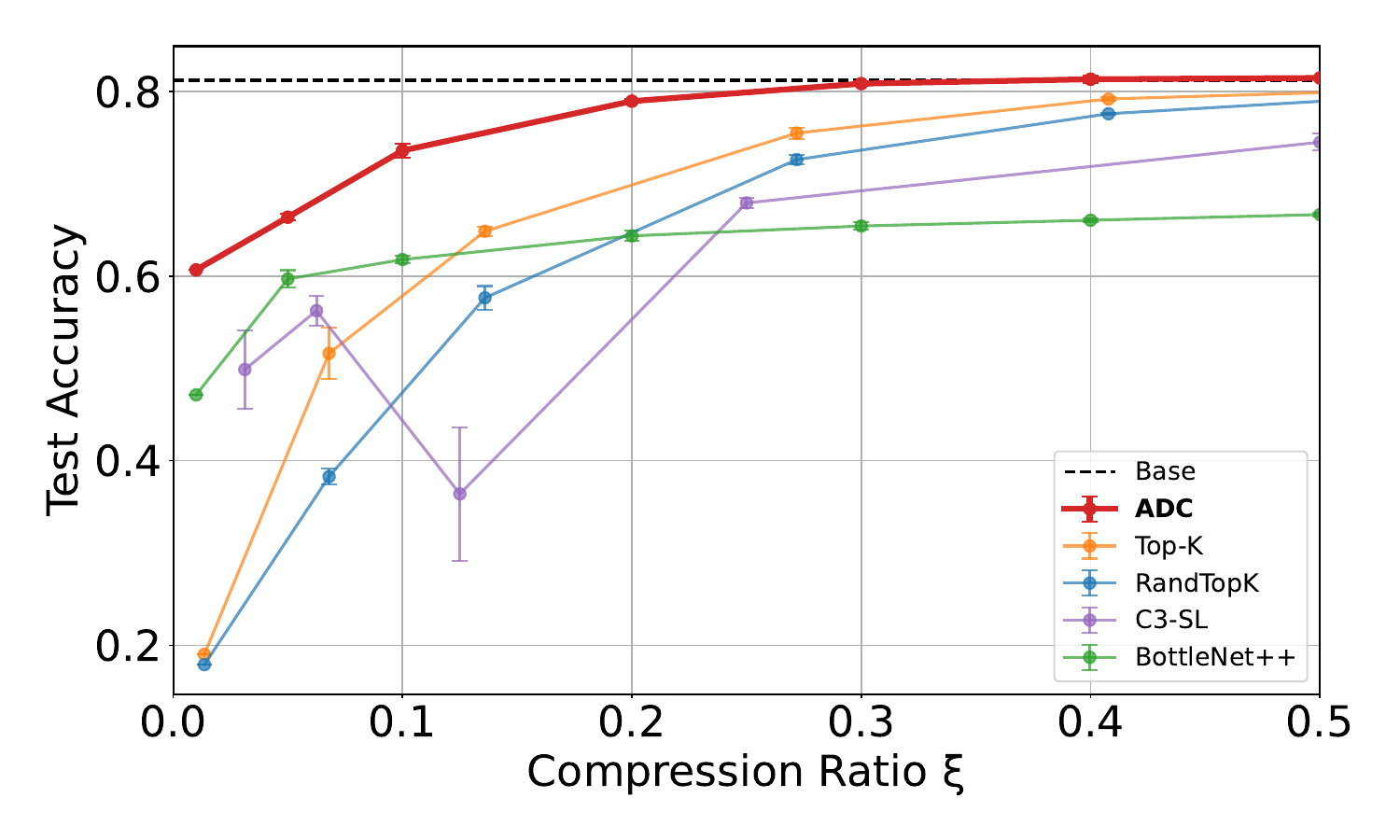}
        \caption{DeiT-T on \food}
        \label{fig:sub3}
    \end{subfigure}
    \hfill
    \begin{subfigure}{0.48\linewidth}
        \centering
        \includegraphics[width=0.92\linewidth]{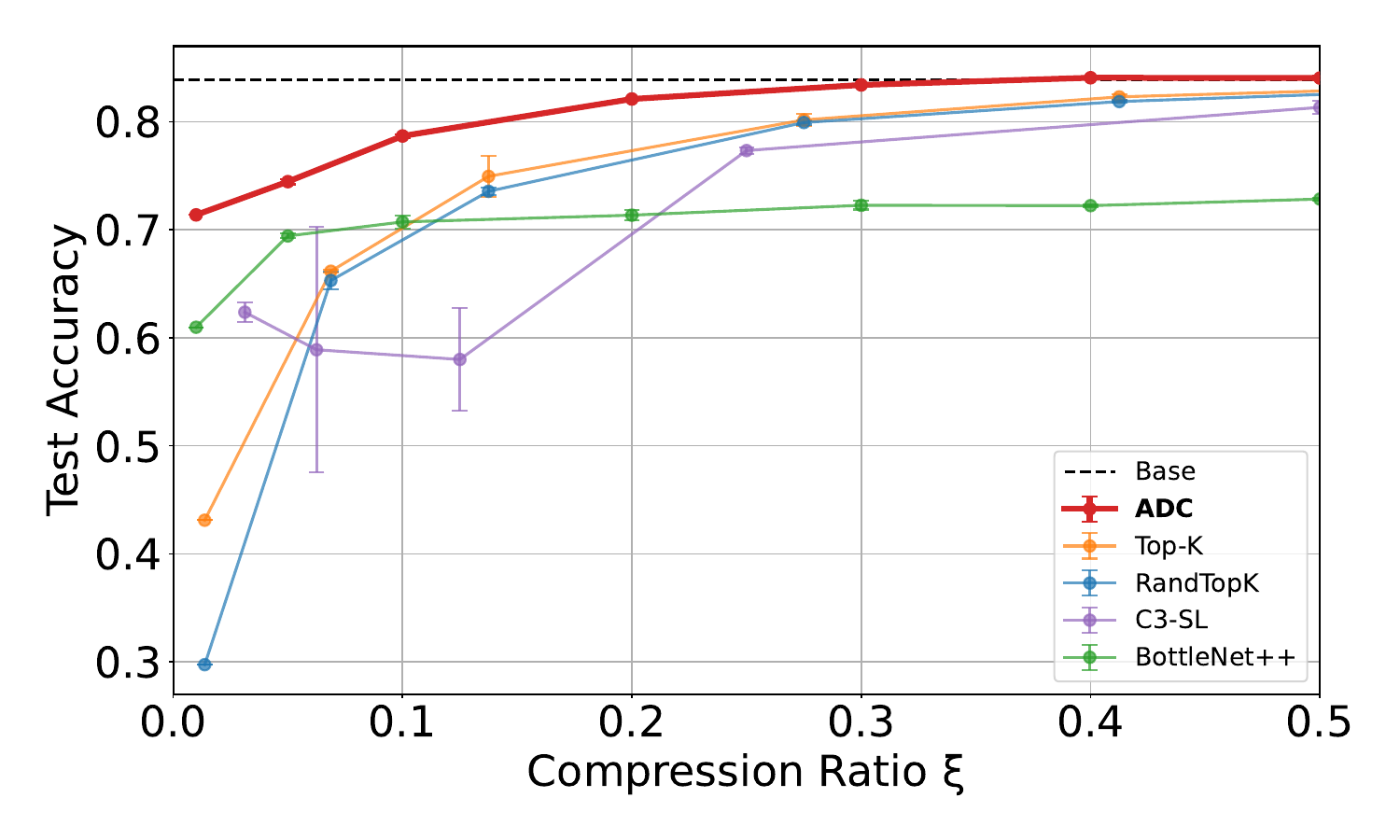}
        \caption{DeiT-S on \food}
        \label{fig:sub4}
    \end{subfigure}

    \caption{Final test accuracy vs compression for each combination of models and datasets.}
    \label{fig:accuracy_vs_compression}
\end{figure*}



    

\begin{figure}[tbp]
    \centering
    \begin{subfigure}{1\linewidth}
        \centering
        \includegraphics[width=0.8\linewidth]{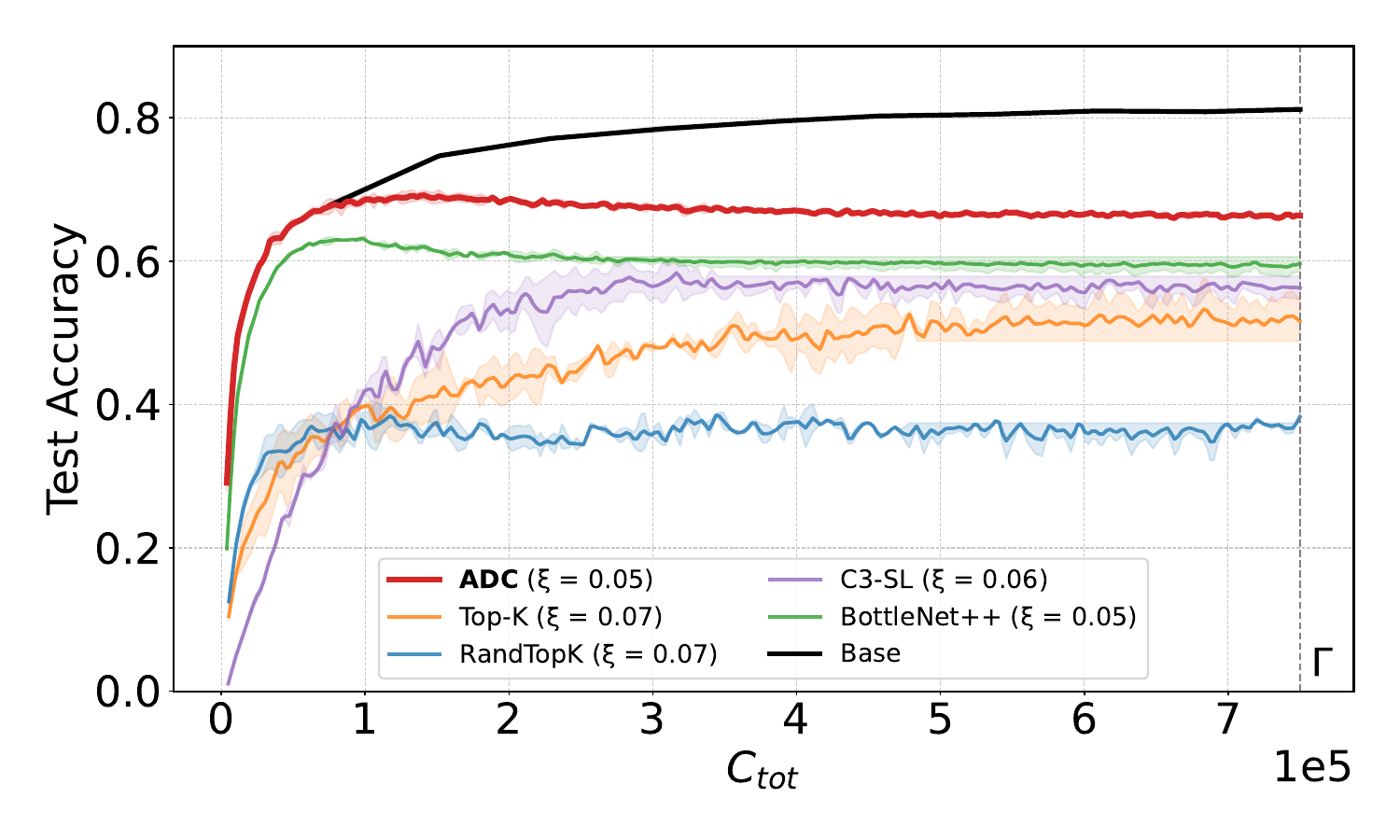}
        \caption{Low $\xi$ Regime}
    \end{subfigure}   
    \\
    \vfill
    \begin{subfigure}{1\linewidth}
        \centering
        \includegraphics[width=0.8\linewidth]{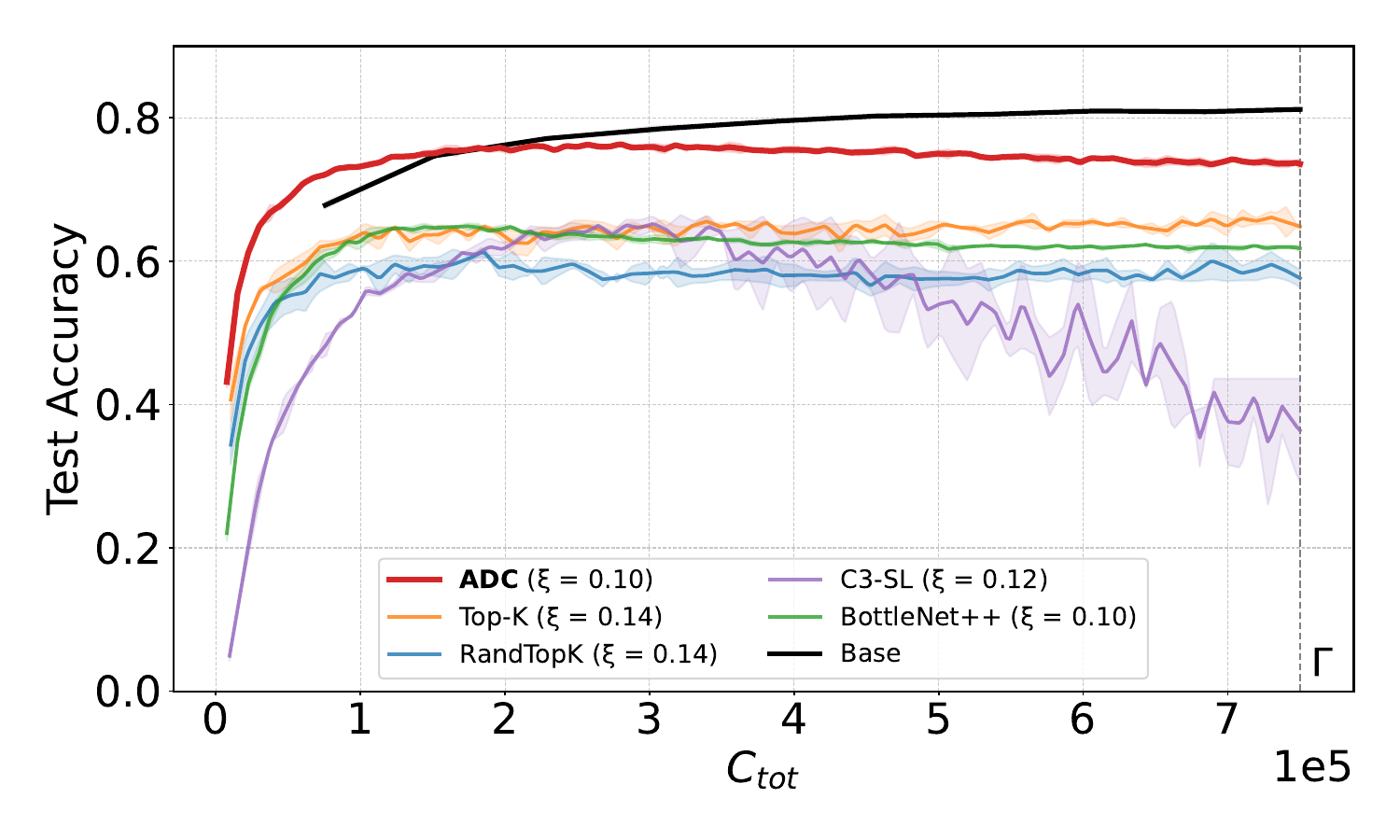}
        \caption{Intermediate $\xi$ Regime}
    \end{subfigure}   
    \\
    \vfill
    \begin{subfigure}{1\linewidth}
        \centering
        \includegraphics[width=0.8\linewidth]{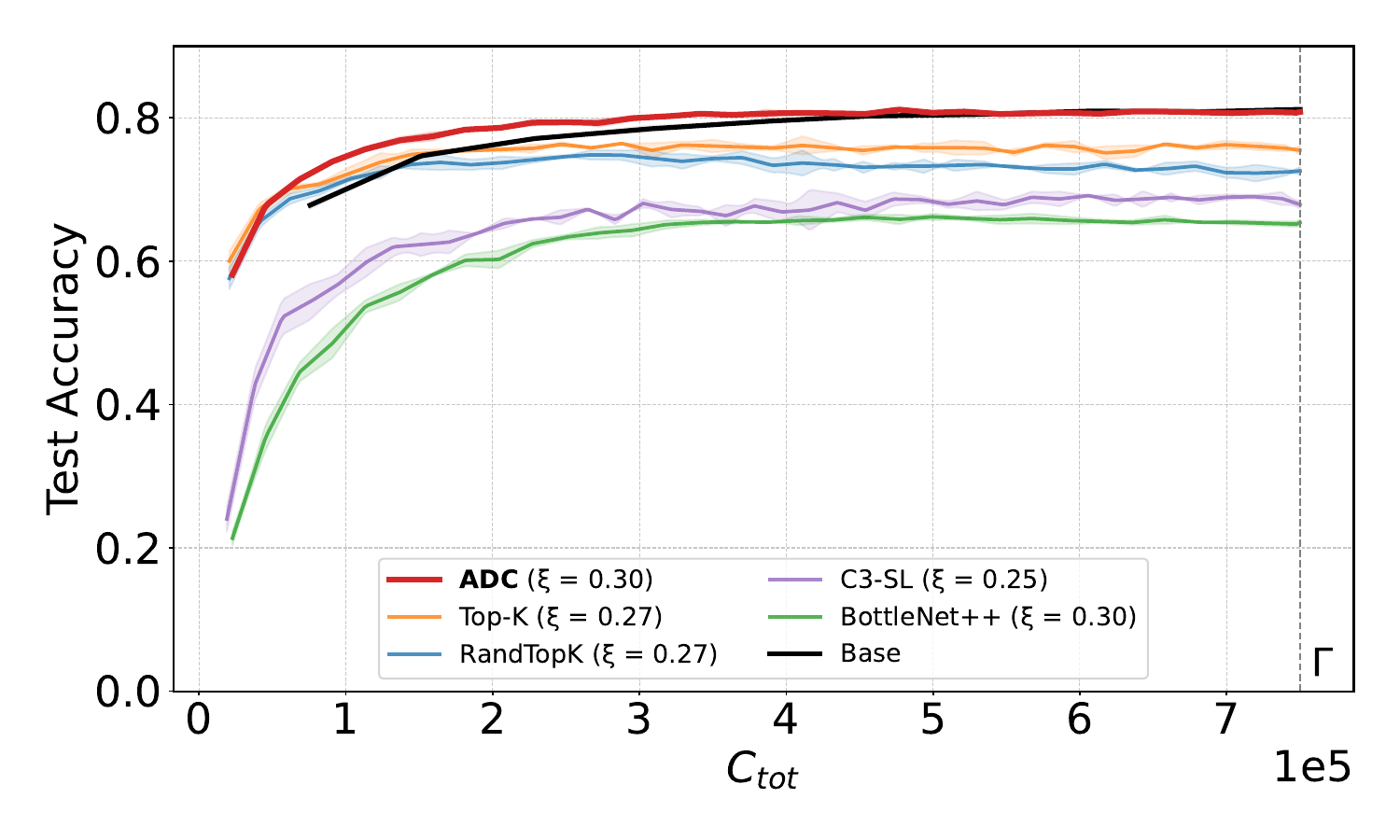}
        \caption{High $\xi$ Regime}
    \end{subfigure}    
    \caption{How test accuracy evolves when training DeiT-T on \food using different methods. Results are shown for different compression ratio regimes.}
    \label{fig:raw_training}
\end{figure}

\Cref{fig:accuracy_vs_compression} reports the test accuracy as a function of the overall compression ratio $\xi$ for all model–dataset pairs, with results averaged over three independent runs. Among the evaluated methods, \methoda is the only approach that consistently preserves high accuracy across both architectures and datasets, achieving good results also when aggressive compression ratios are used.

In particular, \methoda provides the best results in the low-compression ratio regime, where most competing approaches exhibit severe performance degradation. For instance, when training  DeiT-T on \cifar, \methoda achieves near-baseline accuracy already at $\xi \approx 0.1$, whereas alternative methods require substantially higher compression ratios to reach comparable performance.

Top-K and RandTopK remain competitive at moderate to high compression levels, but their performance deteriorates as $\xi$ decreases. C3-SL shows relatively strong behaviour under extreme compression, yet its instability at intermediate values (e.g., $\xi = 0.125$) suggests convergence difficulties during training. BottleNet++ achieves good accuracy at very low compression, but its performance saturates and does not improve as the compression ratio increases.

To further investigate training dynamics, \Cref{fig:raw_training} reports the evolution of test accuracy for DeiT-S on \food across different compression regimes.
Across all settings, \methoda consistently achieves higher accuracy for the same communication cost compared to all baselines, demonstrating superior communication efficiency. The advantage is most pronounced in the low–$\xi$ regime, yet it remains evident in intermediate and high–$\xi$ scenarios as well.
In addition to efficiency, \methoda exhibits remarkably stable convergence, without the sharp fluctuations observed in competing methods (like C3-SL). This stability suggests that our approach not only preserves accuracy under aggressive compression but may also act as an implicit regularizer during training.

\subsection{Ablation experiments}
In this section, we analyse how the components of our proposal affect the final results. To this end, we evaluate the performance when varying the batch size, the splitting point for generating the two networks, and the method used for merging similar activations. 
\subsubsection{Impact of batch size}
%
%
\begin{figure}
    \centering
    \begin{minipage}{1\columnwidth}
    \begin{subfigure}{0.48\columnwidth}
    \centering
    \includegraphics[width=\linewidth]{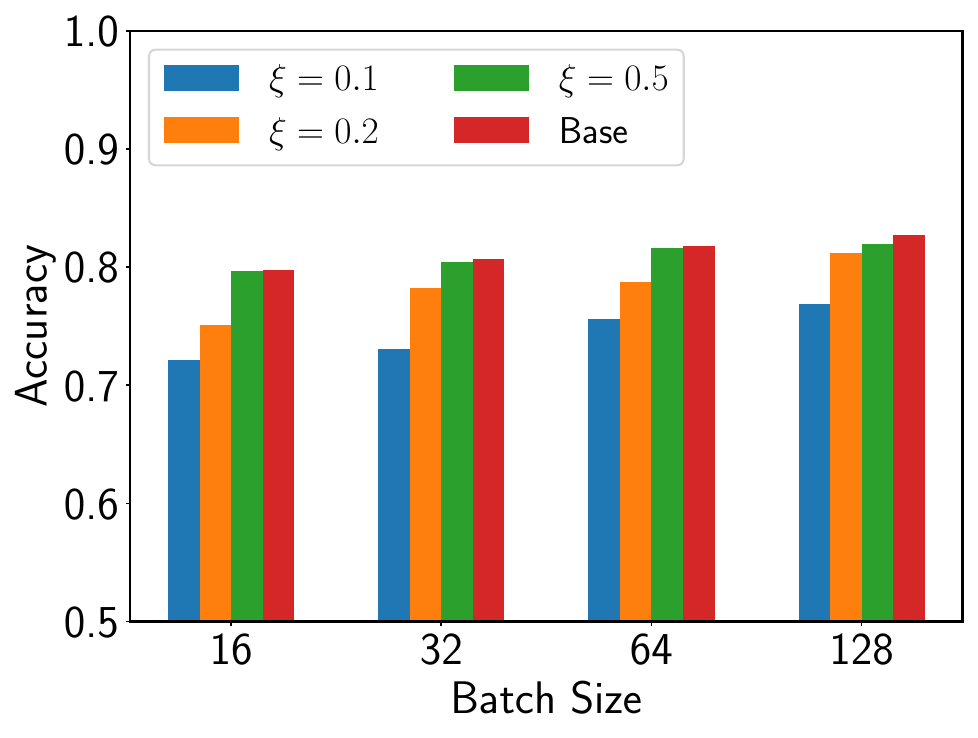}
    \end{subfigure}
    \begin{subfigure}{0.48\columnwidth}
    \centering
    \includegraphics[width=\linewidth]{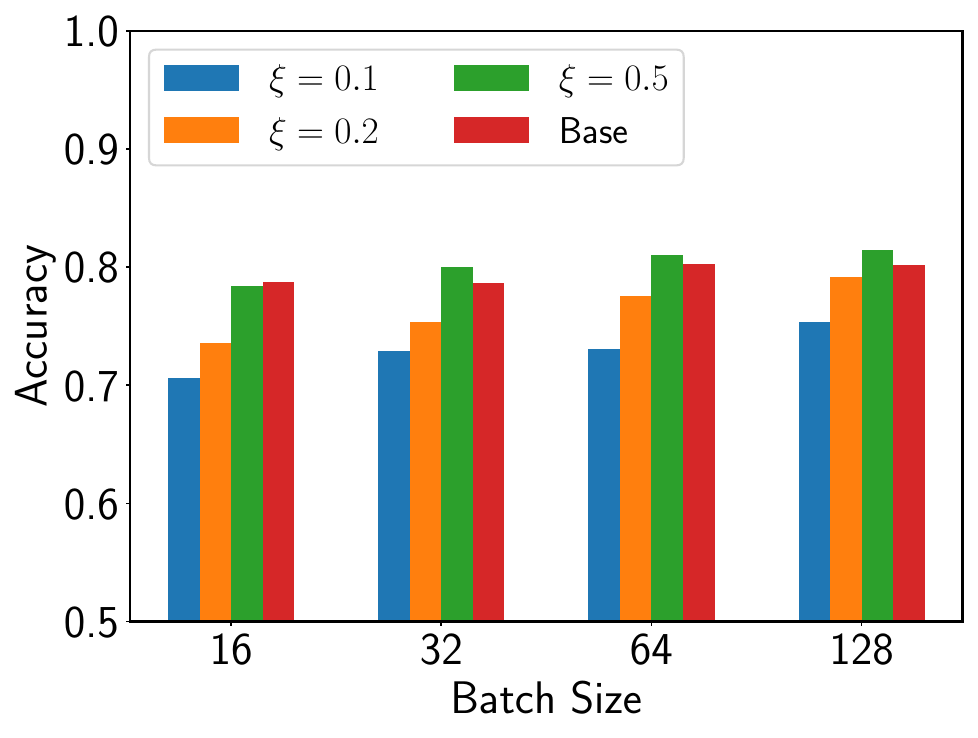}
   
    \end{subfigure}
     \end{minipage}
        
   
    \caption{How the batch size affects the final accuracy of \methoda. The model is DeiT-T, tested with multiple compression ratios on \cifar (left) and \food (right).}
    \label{fig:batch}
\end{figure}

Since our approach merges samples within each batch, analyzing the effect of batch size provides further insight into its behavior. \Cref{fig:batch} reports results across different batch sizes, showing that while performance improves with larger batches, \methoda remains stable even at smaller batch sizes.

We attribute this improvement to the increased diversity of samples within larger batches, which allows the $K$-means to identify clusters with more aligned $\text{CLS}_{score}$ distributions. As a result, the merging process becomes more semantically coherent, and the subsequent token selection—based on the average $\text{CLS}_{score}$ within each cluster—more accurately reflects the most informative token positions.

\subsubsection{Batch merging with different vectors}




%
\begin{table}[t]
\centering
\resizebox{0.8\columnwidth}{!}{%
\begin{tabular}{l|ccc|ccc|}
\cline{2-7}
 & \multicolumn{3}{c|}{\cifar} & \multicolumn{3}{c|}{\food} \\ \cline{1-7} 
 \multicolumn{1}{|c|}{$\xi$} & \multicolumn{1}{c|}{0.05} & \multicolumn{1}{c|}{0.1} & 0.2 & \multicolumn{1}{c|}{0.05} & \multicolumn{1}{c|}{0.1} & 0.2 \\ \hline \hline
\multicolumn{1}{|l|}{$\text{CLS}_{score}$} & \multicolumn{1}{c|}{0.6726} & \multicolumn{1}{c|}{0.7687} & 0.8119 & \multicolumn{1}{c|}{0.6826} & \multicolumn{1}{c|}{0.7369} & 0.7918 \\ \hline
\multicolumn{1}{|l|}{$\text{CLS}_{token}$} & \multicolumn{1}{c|}{0.6529} & \multicolumn{1}{c|}{0.7653} & 0.7964 & \multicolumn{1}{c|}{0.6773} & \multicolumn{1}{c|}{0.7391} & 0.7826 \\ \hline
\multicolumn{1}{|l|}{$\text{AVG}_{token}$} & \multicolumn{1}{c|}{0.6764} & \multicolumn{1}{c|}{0.7687} & 0.7978 & \multicolumn{1}{c|}{0.6640} & \multicolumn{1}{c|}{0.7396} & 0.7824 \\ \hline
\end{tabular}
}
\caption{How the selection of the vector used in batch merging affects the final results. The evaluated model is DeiT-T, trained with different communication constraints $\xi$.}
\label{tab:ablation_pooling}
\end{table}
\begin{figure}[tbp]
    \centering
    \begin{subfigure}{1\linewidth}
        \centering
        \includegraphics[width=0.8\linewidth]{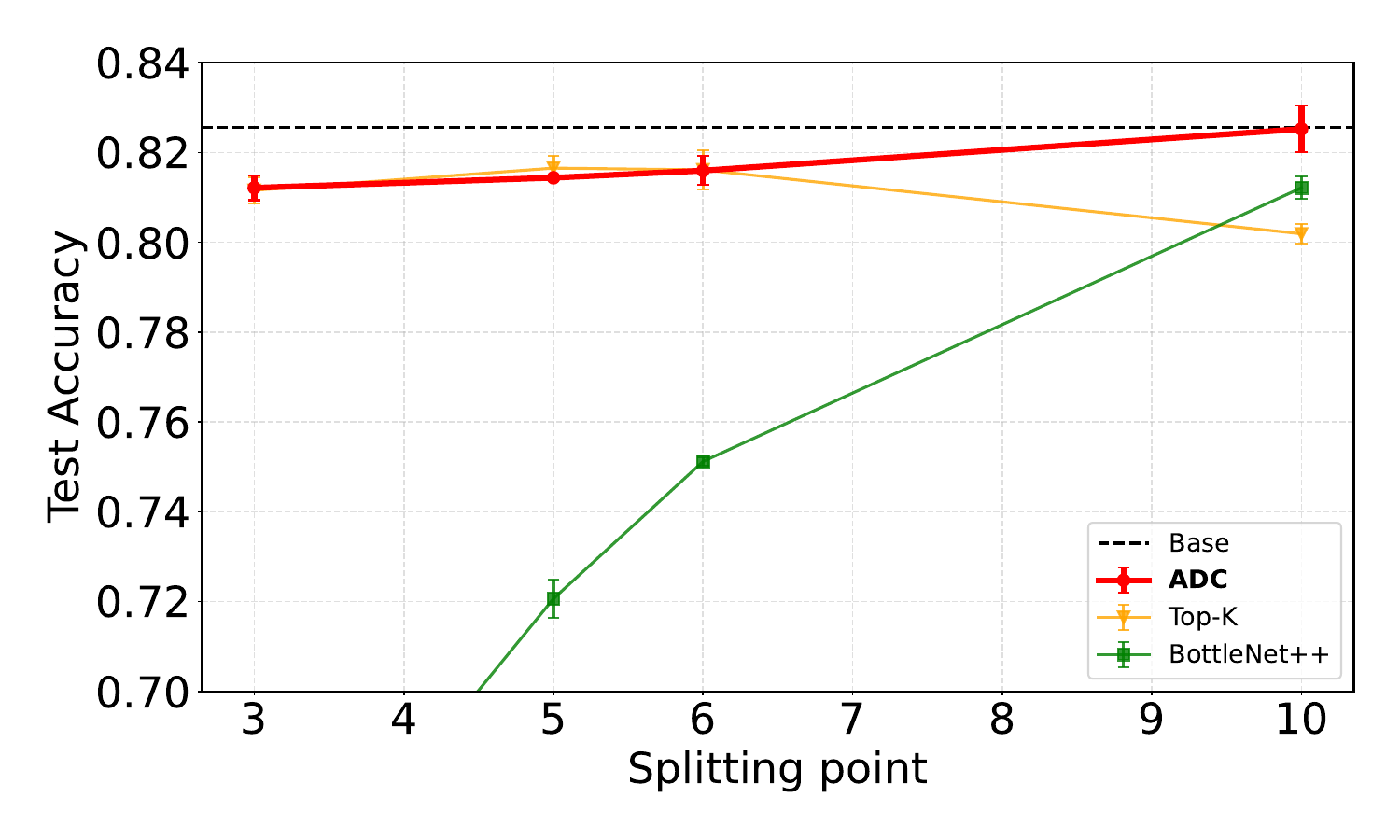}
        \caption{\cifar with $\xi=0.25$}
    \end{subfigure}
    \\
    \vfill

    \begin{subfigure}{1\linewidth}
        \centering
        \includegraphics[width=0.8\linewidth]{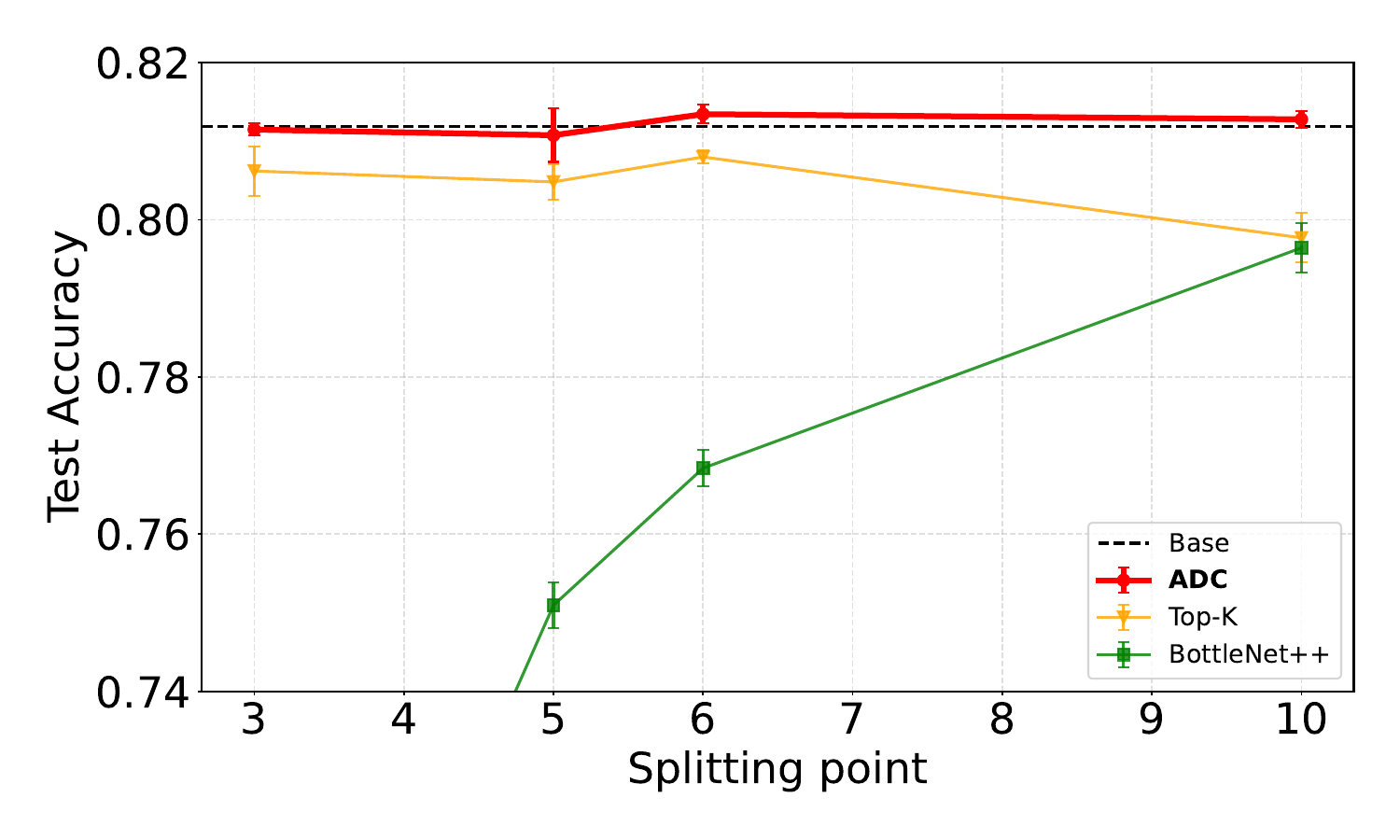}
         \caption{\food with $\xi=0.5$}
    \end{subfigure}    
    \caption{How changing the split point affects DeiT-T training. Some results are omitted for clarity.}
    \label{fig:split-point}
\end{figure}
In this section, we investigate how the choice of the vector used for $K$-means computation during batch merging affects performance. We compare three strategies: the standard $\text{CLS}_{score}$, already introduced in the method section, the $\text{AVG}_{token}$, which merges tokens based on the average of all token activations, and the $\text{CLS}_{token}$, which uses the raw class token for emerging batch samples.

As shown in \Cref{tab:ablation_pooling}, the $\text{CLS}_{score}$ strategy consistently yields the best performance. We attribute this to its alignment with the token selection step, which always relies on the average $\text{CLS}_{score}$ within each cluster to identify the most informative tokens.

When clusters are built using $\text{CLS}_{score}$, the selected tokens naturally reflect shared importance patterns across the samples, preserving semantic consistency. In contrast, alternative strategies may group activations with divergent token importance, leading the averaged vector to emphasize tokens that are only weakly relevant across the cluster. This mismatch results in suboptimal token selection and reduced performance.

%
    

\subsubsection{Shifting the splitting point}

Here, we analyze how changing the splitting point affects the results. The results of such experiments are shown in \Cref{fig:split-point}; it shows the results for DeiT-trained on the two datasets, for two different compression ratios. The results show that our approach improves the results when the splitting point is placed deeper in the model (high splitting point). It happens because in the deeper layer, the class token, used for merging the activations, is more semantically informative. As opposed to Top-K, our approach never loses accuracy. Top-K, instead, loses some accuracy points when the splitting point increases, for the same reason our improves: Top-K discards more informative features the more the splitting point is higher. Bottlenet, instead, always fails with lower splitting points, while recovering when it gets higher, reaching or surpassing Top-K.

\subsection{Activation visualization}
%
    

%

\begin{figure}[tbp]
    \centering

        \begin{subfigure}{0.5\textwidth}
        \centering
        \includegraphics[width=0.9\textwidth]{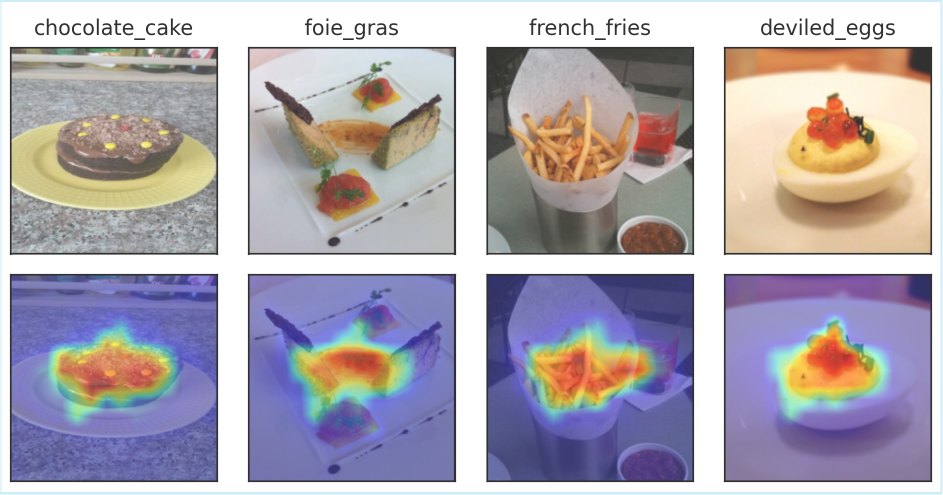}
        \caption{$\xi=0.01$}
        \label{fig:clust2}
    \end{subfigure}
    \vfill
    \begin{subfigure}{0.5\textwidth}
        \centering
        \includegraphics[width=0.69\textwidth]{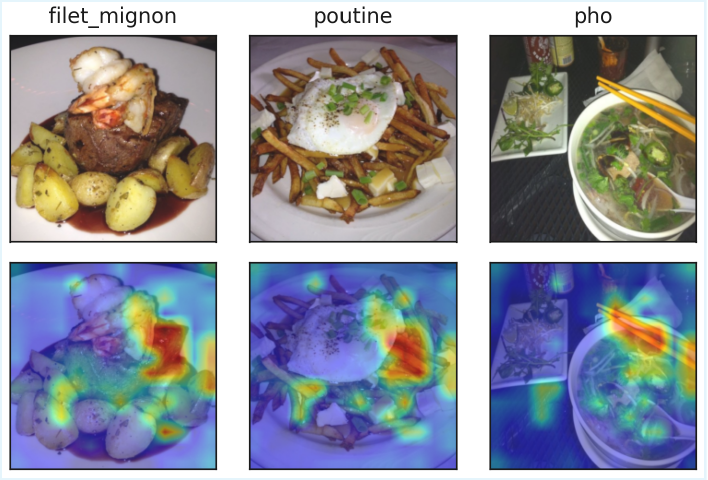}
        \caption{$\xi=0.2$}
        \label{fig:clust1}
    \end{subfigure}
    \\

    \caption{Visualization of clusters obtained with DeiT-T on \food, using different compression ratios. For each cluster, we show the images on top and below the attention rollout \cite{abnar2020quantifying} of the images within the same cluster, considering only the top-k selected tokens. The red regions are the ones with higher attention. }
    \label{fig:merging_heatmaps}
\end{figure}

    In this section, we visually inspect images that are clustered together during the batch merging process. \Cref{fig:merging_heatmaps} presents two representative clusters obtained during the training of DeiT-T on Food-101, at compression ratios of $\xi = 0.01$ \subref{fig:clust2} and $\xi = 0.2$ \subref{fig:clust1}. For each image, we also show the attention rollout \cite{abnar2020quantifying} of the top-$k$ selected tokens.

Despite the images belonging to different classes, we observe consistent attention patterns across both clusters, suggesting that the merged activations remain semantically meaningful. At a lower compression ratio, where only the most relevant tokens are retained, the attention focuses almost exclusively on the primary object within each image. Conversely, when the model is allowed to retain more tokens, the attention becomes more distributed, capturing not only the object but also relevant contextual features, such as the surrounding plate, garnish, or background elements, which may further aid the model's generalization or serve as registers for storing relevant information \cite{darcet2024vision}.
\section{Conclusion and Future Work}

In this work, we introduced {\method}, a novel communication-efficient framework for Split Learning with Vision Transformers. The proposed method leverages a two-stage compression strategy that jointly reduces redundancy across both the batch and token dimensions. By aligning these two forms of compression, {\method} achieves substantially higher compression ratios while preserving model accuracy, even in regimes where all existing baselines fail in achieving high results. This highlights the robustness of our approach and its suitability for deployment in scenarios with stringent communication constraints.

As a future direction, we aim to extend the proposed framework to more realistic communication environments, including noisy and fading wireless channels, where robustness to signal degradation becomes essential. Moreover, applying the framework to multi-client scenarios such as Federated Learning could shed light on how collaborative and distributed training can leverage joint compression across clients, thereby enhancing scalability to large networks of edge devices.

\bibliographystyle{IEEEtran}
\bibliography{main}

\end{document}